\documentclass[letterpaper, 10 pt, conference]{ieeeconf}  

\IEEEoverridecommandlockouts                              

\usepackage{graphicx} 
\usepackage{mathptmx} 
\usepackage{times} 
\usepackage{amsmath} 
\usepackage{amssymb}  
\usepackage{xcolor}
\usepackage{booktabs}
\usepackage{gensymb}
\usepackage{multirow}
\usepackage{hyperref}

\title{\LARGE \bf
Deep-Geometric 6 DoF Localization from a Single Image in Topo-metric Maps
}

\author{Tom Roussel*$^{1}$, Punarjay Chakravarty*$^{2}$, Gaurav Pandey$^{2}$, Tinne Tuytelaars$^{3}$, Luc Van Eycken$^{3}$%
 \thanks{$^1$ Tom is a PhD candidate at KU Leuven, Belgium and worked on this as an intern for Ford.
    {\tt\small tom.roussel@esat.kuleuven.be}}%
 \thanks{$^{2}$ The authors are with Ford Greenfield Labs, Palo Alto, USA.
        {\tt\small pchakra5@ford.com},{\tt\small gpandey2@ford.com}}%
         \thanks{$^{3}$ The authors are at KU Leuven, Belgium.
        {\tt\small \{Tinne.Tuytelaars, Luc.VanEycken\}@esat.kuleuven.be}}
 \thanks{$*$ Equal contribution}
}

\begin{document}

\newcommand{\red}{\color{red}}

\maketitle
\thispagestyle{empty}
\pagestyle{empty}

\begin{abstract}

We describe a Deep-Geometric Localizer that is able to estimate the full 6 Degree of Freedom (DoF) global pose of the camera from a single image in a previously mapped environment. Our map is a topo-metric one, with discrete topological nodes whose 6 DoF poses are known. 
Each topo-node in our map also comprises of a set of points, whose 2D features and 3D locations are stored as part of the mapping process. For the mapping phase, we utilise a stereo camera and a regular stereo visual SLAM pipeline. During the localization phase, we take a single camera image, localize it to a topological node using Deep Learning, and use a geometric algorithm (PnP) on the matched 2D features (and their 3D positions in the topo map) to determine the full 6 DoF globally consistent pose of the camera.
Our method divorces the mapping and the localization algorithms and sensors (stereo and mono), and allows accurate 6 DoF pose estimation in a previously mapped environment using a single camera. With potential VR/AR and localization applications in single camera devices such as mobile phones and drones, our hybrid algorithm compares favourably with the fully Deep-Learning based Pose-Net that regresses pose from a single image in simulated as well as real environments.
\end{abstract}

\section{INTRODUCTION}
We present a method to localize a monocular image stream in a previously mapped environment. In this context, we describe a light-weight system that operates on a single camera image and localizes that camera in a map. Mapping is done once, using a system that has access to depth perception (we use stereo). Subsequently, a second robot - say a drone with a single forward-facing camera - is able to obtain a full 6 Degree of Freedom (DoF) pose along the route taken by the mapping robot.

The philosophy behind this approach is that depth sensing can be demanding - computation, power and expense-wise, and is not always available. Mapping is done once, with a robot that is equipped with depth sensing. After this, robots equipped with monocular sensing are able to localize themselves in this environment, as long as they don't stray too far from the mapped route. This opens up the possibility of small, cheap and low-cost robots, wheeled or aerial, equipped with a single camera, operating in a surveillance or delivery role in a pre-mapped environment.

Our system for monocular localization is hybrid in more ways than one. We use 
both 
Deep Learning and regular geometric techniques that pre-date the revolution in the former and combine topological and metric mapping approaches. 

\begin{figure}
    \centering
    \includegraphics[width=0.95\linewidth]{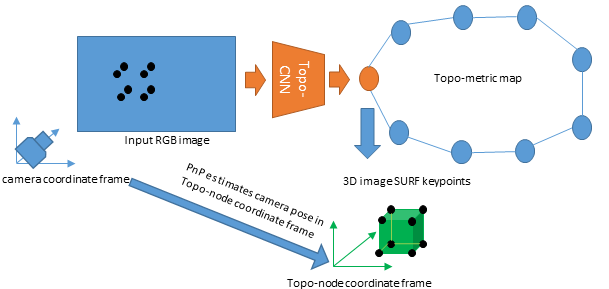}
    \caption{Overview of our Deep-Geometric Localizer. A classifier (such as a CNN) matches the input RGB image to a node in the topo-metric map. The Topo-node stores SURF features for a template image and their 3D coordinates. PnP is used to determine the 6 DoF pose of the camera.}
    \label{fig:topo_metric_overview}
\end{figure}

We present experiments, both simulated and real, where a mapping robot is first driven in a loop through an environment and subsequently, a monocular camera is able to localize itself to a full 6 DoF pose estimate along the same route. The mapping robot uses a stereo camera and runs a state-of-the-art open-source visual SLAM \cite{mur2015orb} algorithm to generate a map of the environment as it traverses the loop.
This map, comprising of a sequence of camera poses and 3D points in the environment is first converted to a lighter topological map, with topological nodes at periodic intervals along the path taken by the robot. Each topo-node comprises of an image, 2D feature points (along with their visual descriptors), and their corresponding 3D locations. The mapping robot traverses the same route multiple times to build up a database of images for each topo-node. 
We train a deep learning based-classifier to classify images from the same route as belonging to one of the topo-nodes. In this work, we take a look at 3 different classifiers: an end-to-end trained classification CNN, a K-nearest neighbour classifier (KNN) using CNN features and a KNN using a NetVLAD feature descriptor~\cite{arandjelovic_netvlad}. 
This allows a robot with a monocular camera to use a single image to determine the closest topo-node. 2D features are then matched between the new image and the features in the database for this node (whose 3D positions are known), followed by Perspective-n-Point (PnP) projection \cite{lepetit2009epnp} to determine the pose of the robot to be localized. We also use a patch-normalization pre-processing step for making the feature matching invariant to lighting change. Thus, we use a hybrid of conventional geometric techniques and recent advances in Deep Learning, to localize a camera in a previously mapped environment. The system is illustrated in Figure \ref{fig:topo_metric_overview}.

\section{RELATED WORK}
\subsection{Visual SLAM - SfM and VO}
The goal of Visual SLAM, or Visual Simultaneous Localisation and Mapping, is to use a moving camera to map an environment. The camera needs to simultaneously determine the 3D geometry of the structure in the scene and estimate its own moving position in this incrementally mapped structure. The structure estimation and camera trajectory estimation sub-problems are sometimes studied independently, depending on the relative importance of the two, and called Structure from Motion (SfM) and Visual Odometry (VO) respectively. 

A typical SfM formulation \cite{agarwal2009building} determines the 3D structure in the scene from a set of (not necessarily sequential) images. Image features are matched over multiple viewpoints and camera poses and the 3D geometry are jointly optimized over all the cameras. An additional texture mapping step ensures visual realism.

In contrast, VO emphasizes the accuracy of camera trajectory estimation that often needs to be real-time for robotics applications. Therefore, in case of VO, the 3D structural mapping is generally limited to the estimation of 3D point locations which are backprojected into the scene from their 2D image counterparts (using stereo or parallax from camera motion in the monocular case), and the optimization of the sequence of camera poses and the 3D point estimates are done incrementally in real-time. A comprehensive tutorial for VO can be found in \cite{fraundorfer2011visual}.

ORB-SLAM \cite{mur2015orb} is a modern real-time open-source visual SLAM algorithm. As the name suggests, ORB features
are extracted from images. These 2D feature points are then backprojected into 3D. The pose of the camera with respect to these points is determined by Perspective-n-Point (PnP) projection that optimizes the pose based on minimizing the reprojection error between the 2D projection of 3D points and their tracked counterparts obtained from optical flow across images. 3D points and pose information are stored for key-frames and a backend optimization procedure \cite{kummerle2011g} operating in a separate thread is used to jointly optimize all 3D point and pose estimates when loop closure is detected. 
ORB-SLAM can be used with both monocular and stereo sensors. The former is less accurate and can only map the environment up to an arbitrary scale.

\subsection{Topo-metric Localization}

Topological maps are a discretization of a continuous metric environment into nodes along paths frequented by users of the map.
The topology of these nodes is described by edges between them that encode their geometric or physical arrangement. 
Localization in a topological map involves a determination of the node in the map the robot is currently at. In contrast, metric localization involves determining its precise metric $x,y,z,roll,pitch,yaw$ 
location in the map. Topological representations are more compact, yet less precise compared to metric maps. A hybrid approach involves having a local metric map for each topological node and is a good compromise between the two.
\cite{tomatis2003hybrid} is an early example of combining topological and metric maps. They discretize the environment into topological nodes, with local metric maps at each topo-node. A planar LIDAR is used to determine straight-line structures in the environment and topological nodes are automatically constructed at corridor junctions using corner detection. 
\cite{murillo2007surf,badino2011visual} describe topo-metric localization using a camera. In \cite{badino2011visual}, a vehicle equipped with GPS and a camera is used to make a topological map with an image associated with each topological node. During localization, image matching is used to determine the location of the vehicle closest to the nearest topological node. 
\cite{murillo2007surf} use an omni-directional camera for localization in a topo-metric map, where the bearing of visual landmarks across 3 locations are used to triangulate the position of the camera. Like \cite{murillo2007surf,badino2011visual}, we also use a hybrid topo-metric map, but in contrast to them, we describe full 6 DoF pose estimation of a single image from a perspective camera. \cite{dayoub2013vision} is perhaps the closest conceptually, to our work in that they also use visual place recognition to localize to a topological node, followed by 6 DoF pose estimation by matching to a local point cloud. However, they use a stereo camera throughout, and their pose estimation is obtained by 3D-to-3D matching, whereas we use 3D to 2D matching and are able to localize using a monocular sensor.

Modern visual SLAM approaches like ORB-SLAM \cite{mur2015orb} described earlier implicitly use a topo-metric representation of the environment. Each keyframe is stored as a node in the graph and edges between nodes describe the SE(3) pose transformation between successive keyframes. We use ORB-SLAM, with stereo images, to generate our topometric map. The nodes in our map are much sparser and more distant compared to the keyframes in the ORB-SLAM map. And instead of the ORB features to match images, we treat the image matching to each node as a classification problem and train a Deep CNN to do this.

\subsection{Deep Learning for monocular 6 DoF pose estimation} 
With the recent success of Deep Learning, there have been efforts \cite{kendall2015posenet,kendall2017geometric,li2018undeepvo} at treating pose estimation from a single image as an end-to-end learning problem.
The recent Posenet \cite{kendall2015posenet} and its extension \cite{kendall2017geometric} (with a geometric loss function) use Deep Learning to regress 6 DoF global pose of the camera from a single image. Training poses are provided from an SfM pipeline. 
Undeep VO \cite{li2018undeepvo} also uses Deep Learning to train separate networks for VO and for depth estimation from a single image. They use a stereo camera to train their networks with pose and photometric losses between the left and right images, and during test time, their networks are able to take a monocular image stream from the same environment and give depth maps and a VO output. However, unlike us, their VO pose output is a relative pose between the frames, and not an absolute global pose with respect to a prior map and therefore cannot be used for global localization. 

\section{METHOD}
The following sections describe our system in more detail. We use a stereo camera and a visual SLAM algorithm to generate the topo-metric map. Localization is done in a coarse-to-fine Deep-Geometric approach that first localizes the image to a topo-node and then uses geometry to determine its fine-grained 6 DoF pose.

\subsection{Constructing the Topo-metric map}

We use a stereo camera (stereo ZED \cite{stereozed}) and the open-source visual SLAM algorithm, ORB-SLAM2 \cite{mur2015orb} to extract the camera trajectory, i.e.~a sequence of 6 DoF poses of the moving stereo pair. This trajectory is then discretized to a set of topological nodes in our topo-metric map. We found that SURF features facilitate much higher quality matching, so we elected to use that instead of ORB features. Each node comprises of a single template node image, its global SE(3) pose and a point cloud comprising of 2D SURF features \cite{bay2006surf} and their 3D positions. 
The global SE(3) pose of each node is also stored in the map. Nodes are created based on heuristically determined translation and rotation parameters. These can be 2-20m apart, depending on the size of the environment.
A new node is created when the following condition holds:

\begin{equation}
    D_{total} = D_{translation} + \lambda \cdot D_{rot} > D_{thresh}
    \label{eq:thresh}
\end{equation}
Where $D_{translation}$ is the Euclidean distance the camera has moved and $D_{rot}$ is the angular distance in radians. $\lambda$ is a constant that determines how much rotations contribute the creation of new nodes. $D_{thresh}$ is the threshold distance.

The images in the vicinity of each topo-node are used to train a classifier that can classify RGB images to topo-nodes. 
We consider 3 different classifiers for this task. The first one is a more naive approach, we use a VGG-11 network \cite{simonyan2014very}, pretrained on ImageNet, and replace its output layer with a fully-connected layer with the same number of outputs as nodes in our topo-metric map. We then train the network to directly estimate the nearest topological node for each image in our mapping sequences. We refer to this setup as \textit{CNN-FC}.

Our second classifier utilizes a pretrained VGG-16 network as a feature extractor. For all images in our mapping sequence, we extract a feature vector using the pretrained network and use it to construct a K-nearest neighbour classifier. The network is not retrained. We refer to this setup as \textit{CNN-NN}.

The last classifier we consider is a VGG-16 network, with the final layers replaced with a NetVLAD layer~\cite{arandjelovic_netvlad}. Like the last classifier, we build a K-nearest neighbour classifier using the feature vector computed by the NetVLAD architecture.
We use pretrained weights for the network, which have been trained on the Pittsburgh 30K dataset\footnote{Code and weights taken from \url{https://github.com/Nanne/pytorch-NetVlad}}. 
This architecture has been shown to be able to deal with challenging situations such as lighting changes and large time differences. We refer to this setup as \textit{VLAD-NN}.

\subsection{Deep-Geometric Localization}
Our localizer is a hybrid between newer Deep Learning techniques and geometric computer vision methods.
We use a classifier (as described previously) to localize a single camera image to one of the topological nodes in the map. SURF keypoints \cite{bay2006surf} in the test camera image are matched to keypoints from the template image of the topo-node using a fast nearest neighbour-based matcher~\cite{muja2009fast}. Next, the geometric Perspective-n-Point (PnP) algorithm
is used to determine its SE(3) pose with respect to the topo-node. The PnP algorithm takes in the set of 3D points in the reference frame of the topo-node, and their 2D projections in the current camera image plane (for the matched SURF features), and outputs the pose of the camera in that reference frame. We utilize the Ransac-ed version of PnP from OpenCV, that applies the non-linear Levenberg Marquardt \cite{levenberg1944method} optimization algorithm to find the best pose of the camera (the intrinsics of the camera are assumed known) given matching feature points. Once the pose of the camera is known with respect to the topo-node, and given the global pose of the topo-node found during the mapping process, the global 6 DoF pose of the camera image is easily obtained using a sequential concatenation of SE(3) poses. 

The PnP solver is not bounded, so occasionally when keypoint matching fails, the pose will have jumped by a large distance. These large jumps are easily detected by checking if the distance travelled between two frames is not larger than a threshold (set between 1-5m, dependent on the scale of the environment). When such a jump is found, we copy the previous pose from the trajectory, ignoring the incorrect localization found by the PnP solver.
In the future, this can be replaced by a more sophisticated mechanism that uses the current velocity of the camera.

\subsection{Patch normalization for Robustness to Lighting Change}

\begin{figure}
    \vspace{0.1cm}
    \centering
    \includegraphics[width=0.8\linewidth]{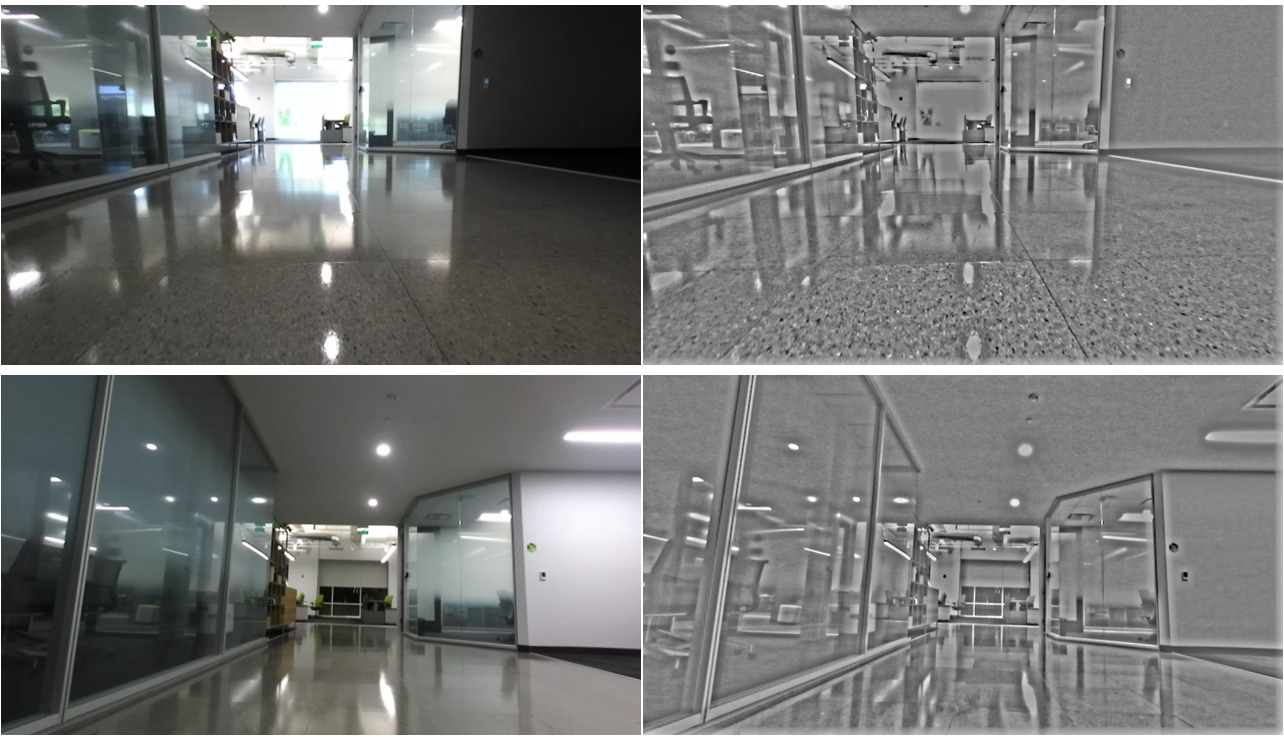}
    \caption{Effect of patch normalization on two images from the same topo-node in the Corridor dataset, 
    taken at different times (day vs night, lights off vs on). Their patch-normalized versions are shown along-side. Patch-normalization improves the feature matching across lighting changes.}
    \label{fig:patch_norm}
    \vspace{-0.5cm}
\end{figure}

We found that the SURF feature matching is more robust across lighting change if we patch-normalized~\cite{zhang2007robust} the images as a pre-processing step. Instead of finding normalizing parameters across the entire image, we do this in 17x17 windows according to the following formula: 
\begin{equation}
    N(p) = \frac{I(p) - \mu_P }{\sigma_P + \epsilon} 
\end{equation}
Where: $N$ is the normalized image, $p$ is the centre of the patch, $\mu_P$ is the mean of the current patch, $\sigma_P$ is the standard deviation of this patch and $\epsilon$ is a small value to prevent division by zero. Figure \ref{fig:patch_norm} shows the effect of patch-normalization across time in the Corridor dataset. This is from the same topo-node in the map, but the images are taken during the day (when outdoor lighting streams in from large windows) and during the night (when the windows are dark, but indoor lights have come on). This pre-processing step is important for images in areas with significant lighting changes, as it results in more robust feature matching.

During preliminary experiments, we found that applying patch normalization before computing the SURF keypoints improved the localization accuracy by $\sim 20\%$ on non-simulated data.

\section{EXPERIMENTS}
\subsection{Simulated Environments}
In an effort to make our results more reproducible, we first verify our method using the open-source autonomous driving simulator, CARLA~\cite{Dosovitskiy17}. It allows us to simulate camera trajectories and collect photo-realistic outdoor RGB images and dense depth maps, along with accurate ground truth camera poses. 
We record sequences in this environment by selecting a path and moving a camera along that path. The approximate size of this environment is 333m x 81m.
We consider 2 kinds of changes in the environment, creating different experiments. In the first experiment, we apply an offset to the trajectory, so the camera is moved further to the right. We split the data into mapping and testing sequences.  This split is chosen such that the sequences are in different lanes on the road. We show the results of this experiment in table \ref{tab:localizer_comparison}.

\begin{figure}
    \centering
    \includegraphics[width=1\linewidth]{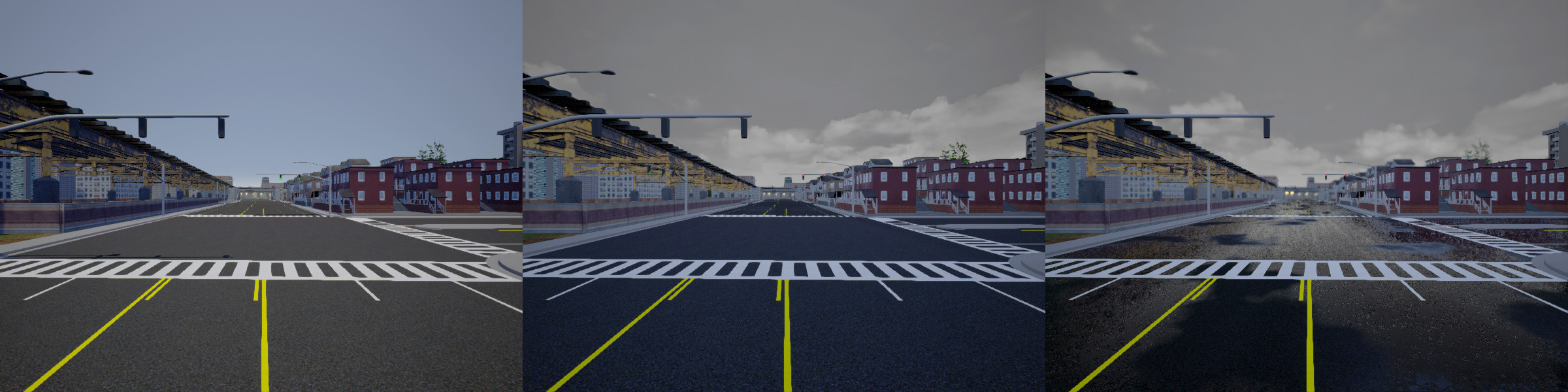}
    \caption{Different weather conditions in CARLA}
    \label{fig:carla_weather}
\end{figure}

To test how well our methods can deal with changes in the environment, we generate sequences with different weather conditions. We use the same mapping sequence as our previous experiment, but we attempt to localize in a cloudy and rainy environment. We show the visual difference between the mapping and localizing sequences in figure \ref{fig:carla_weather}. We show the performance decreases across our proposed method in table \ref{tab:weathertable} we show the topological accuracy (the classification accuracy of our coarse localizers) in figure \ref{fig:topo_acc}. 

\begin{figure}
    \vspace{0.2cm}
    \centering
    \includegraphics[width=0.95\linewidth]{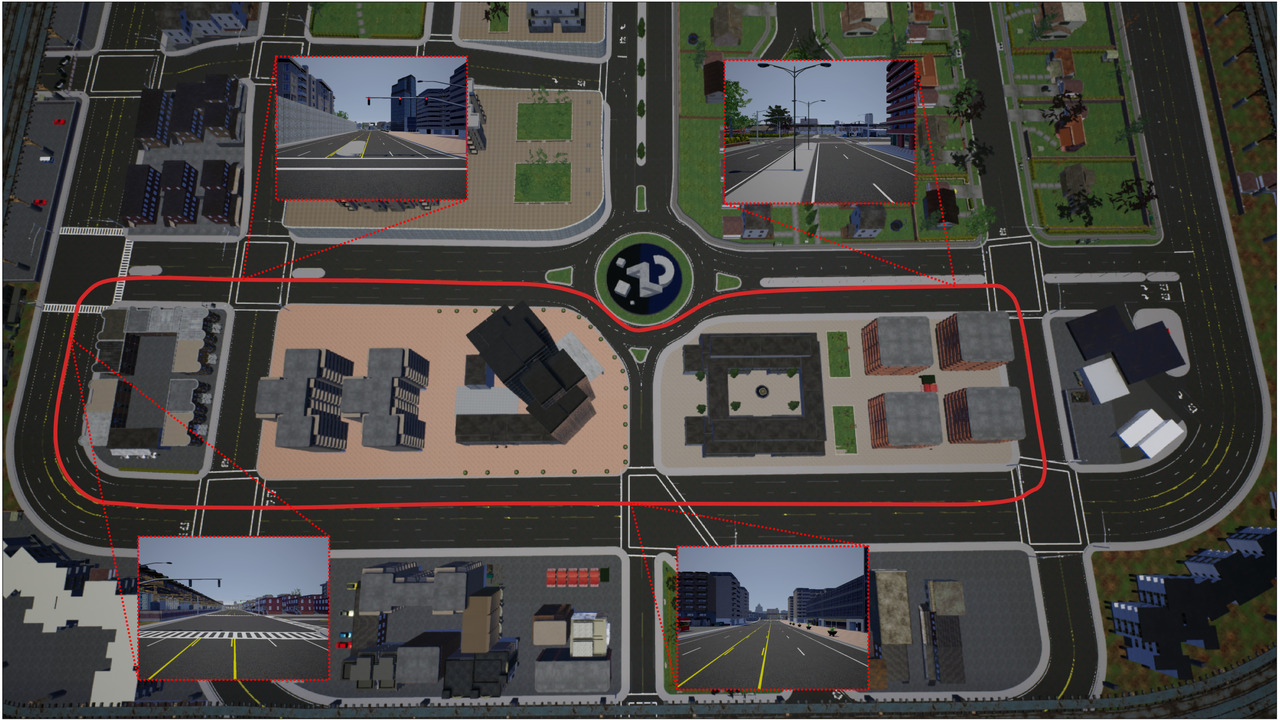}
    \caption{A top down view of the trajectory followed in the CARLA simulator with insets showing camera views along the trajectory}
    \label{fig:carla_top}
\end{figure}

For all our experiments with this dataset, we use $D_{thresh}=20$ and $\lambda=2$.
In figure~\ref{fig:carla_top}, we show the trajectory followed along with some example images from the simulator.

\begin{figure}
    \centering
    \includegraphics[width=0.8\linewidth]{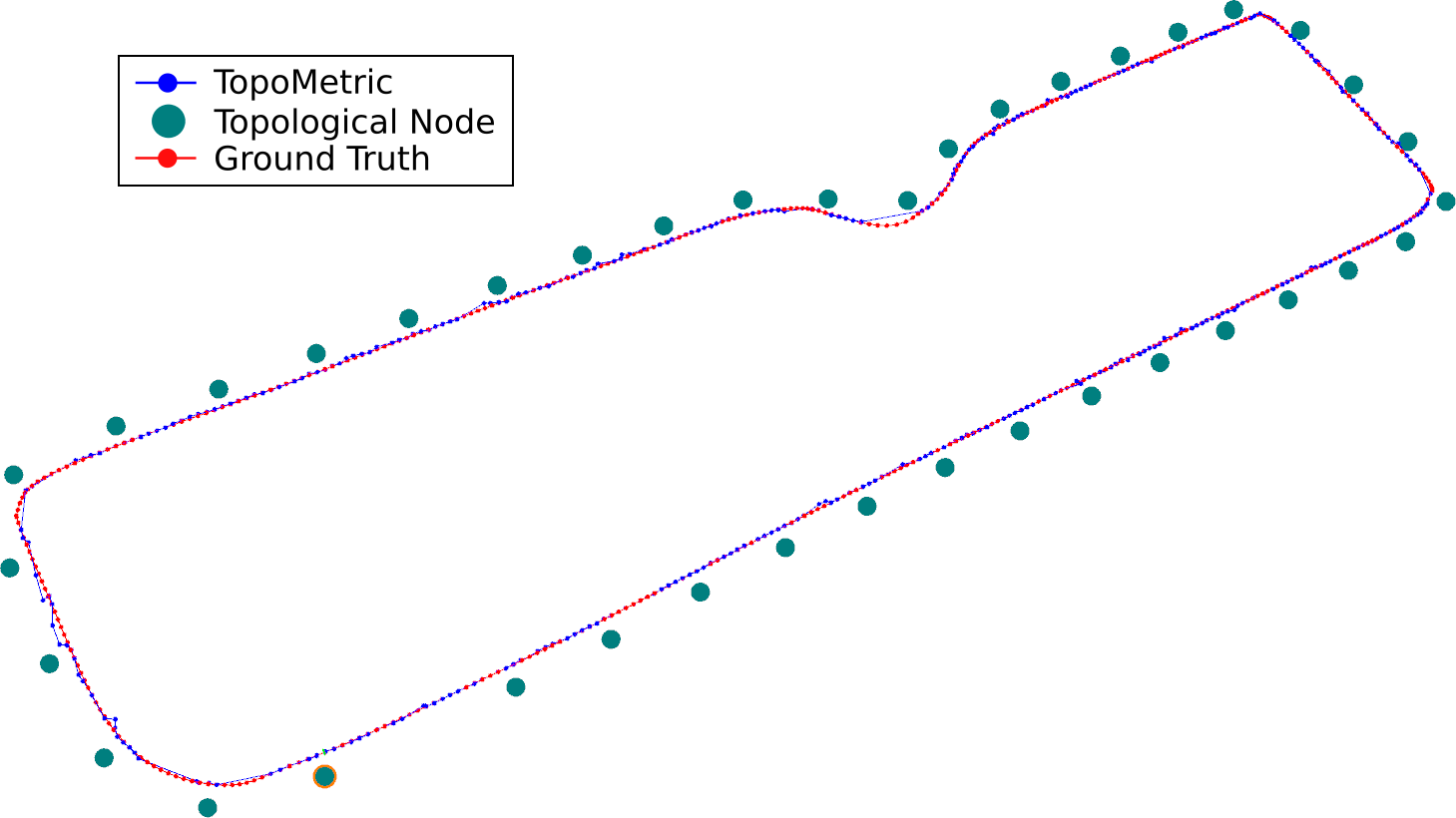}
    \caption{Example localization within the CARLA environment. The conditions shown in the lefmost image are used in the mapping sequences.}
    \label{fig:carla_pangolin}
\end{figure}


\begin{table}[]
\caption{Median translation and rotation errors of posenet and our method}
\centering
\label{tab:localizer_comparison}
\begin{tabular}{@{}ccccc@{}}
\toprule
                          & Posenet & CNN - FC & CNN - NN & VLAD - NN \\ \midrule
\multirow{2}{*}{CARLA}    & 3.114 m   & 1.842 m    & \textbf{1.822 m}    & 1.845 m     \\
                          & 0.810\degree   & 0.604\degree    & \textbf{0.601\degree}    & 0.644\degree     \\ \midrule
\multirow{2}{*}{Corridor} & 0.529 m   & 0.0458 m   & 0.0451 m   &\textbf{ 0.0446 m}    \\
                          & 3.377\degree   & 0.945\degree    & 0.943\degree    & \textbf{0.937\degree}     \\ \midrule
\multirow{2}{*}{Lab}      & 0.131 m   & 0.0207 m   & \textbf{0.0202 m }  & \textbf{0.0202 m}    \\
                          & 1.015\degree   & 0.328\degree    & \textbf{0.350\degree}    & {0.363\degree}    \\ \bottomrule
\end{tabular}
\end{table}


\begin{table}[]
\centering
\caption{Performance decreases for significant weather changes in the environment. CARLA Clear contains the weather seen during mapping. }
\label{tab:weathertable}
\begin{tabular}{@{}cccc@{}}
\toprule
                                  & CNN - FC & CNN - NN & VLAD - NN \\ \midrule
\multirow{2}{*}{Baseline - CARLA} & 1.845 m    & 1.842 m    & \textbf{1.822 m}     \\
                                  & \textbf{0.601\degree}    & 0.644\degree    & 0.604\degree     \\ \midrule
\multirow{2}{*}{CARLA cloudy}     & 2.526 m    & 2.702 m    & \textbf{2.371 m}     \\
                                  & \textbf{0.806\degree}    & 0.745\degree    & 0.811\degree     \\ \midrule
\multirow{2}{*}{CARLA rain}       & 2.982 m    & 2.817 m    & \textbf{2.812 m }    \\
                                  & 0.795\degree    & 0.857\degree    & \textbf{0.787\degree}     \\ \bottomrule

\end{tabular}
\end{table}

\subsection{Corridor \& Lab environments}

\begin{figure}
    \centering
    \includegraphics[width=0.8\linewidth]{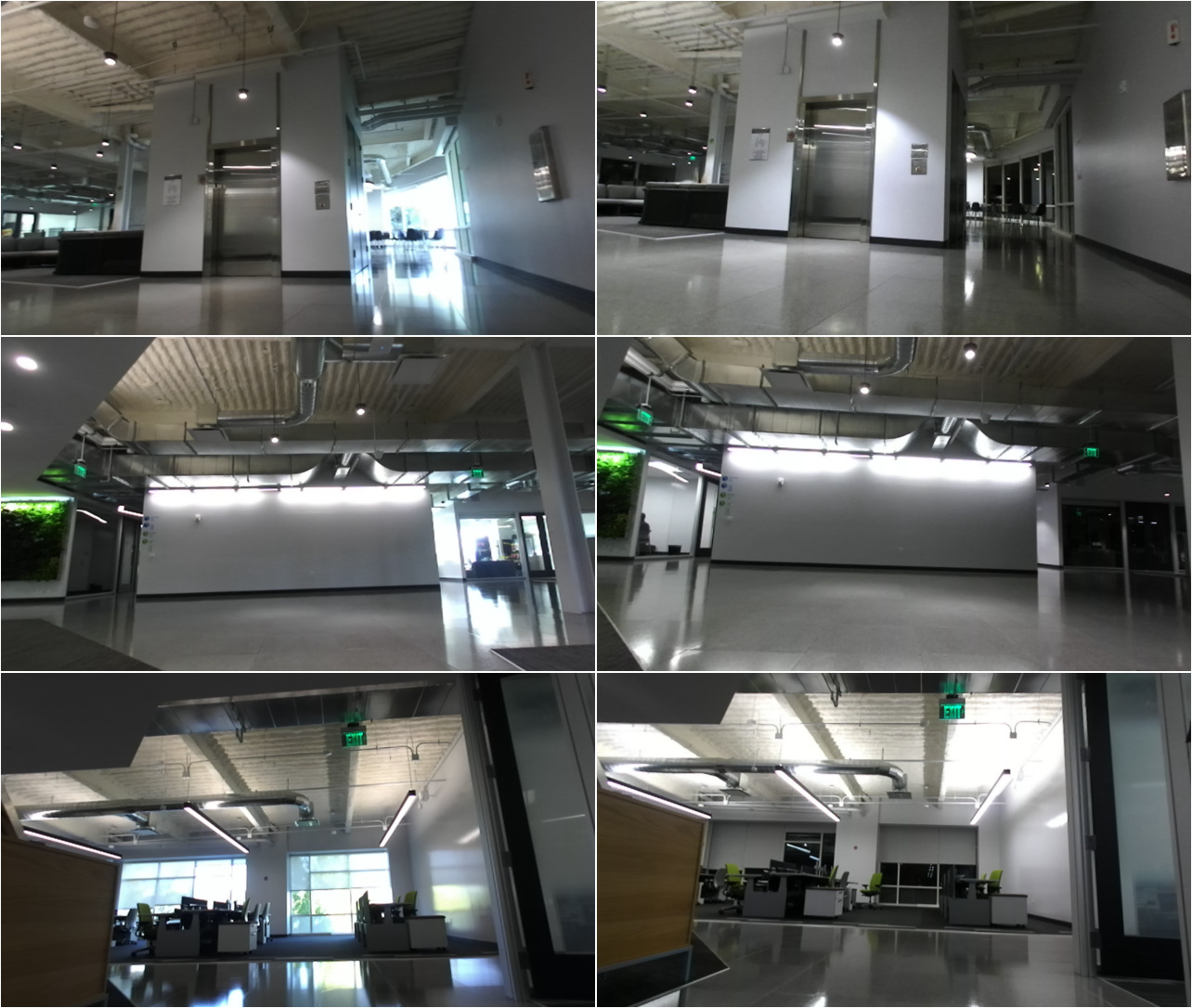}
    \caption{Lighting differences between mapping and localization sequences. The images on the left were used to map the environment, while the images on the right were used to test the localization.}
    \label{fig:light_corr}
\end{figure}
\begin{figure}
    \centering
    \includegraphics[width=0.8\linewidth]{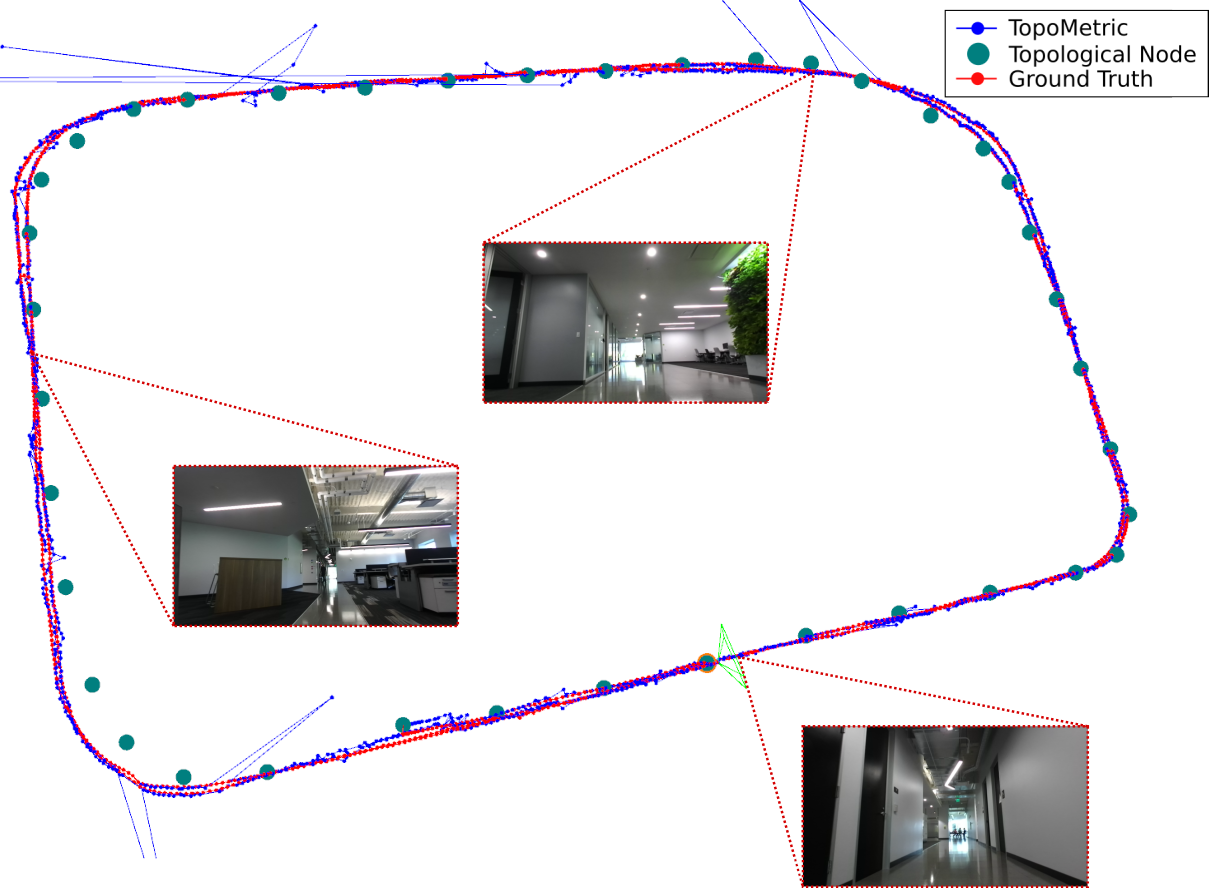}
    \caption{Example localization within the corridor environment with insets showing the different views in this environment.}
    \label{fig:cor_pangolin}
\end{figure}

\begin{figure}
    \centering
    \includegraphics[width=0.6\linewidth]{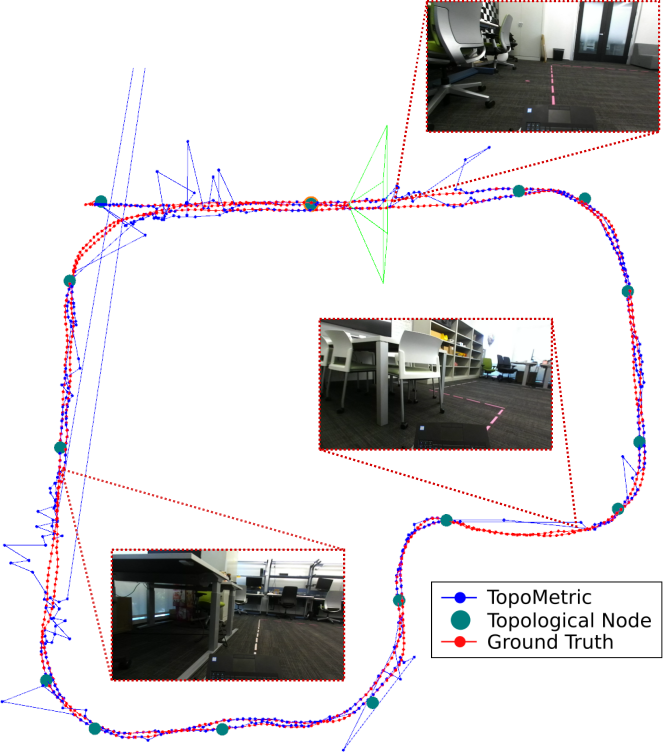}
    \caption{Example localization within the lab environment with insets showing the different views in this environment. }
    \label{fig:lab_pangolin}
\end{figure}

\begin{figure}
    \centering
    \includegraphics[width=0.75\linewidth]{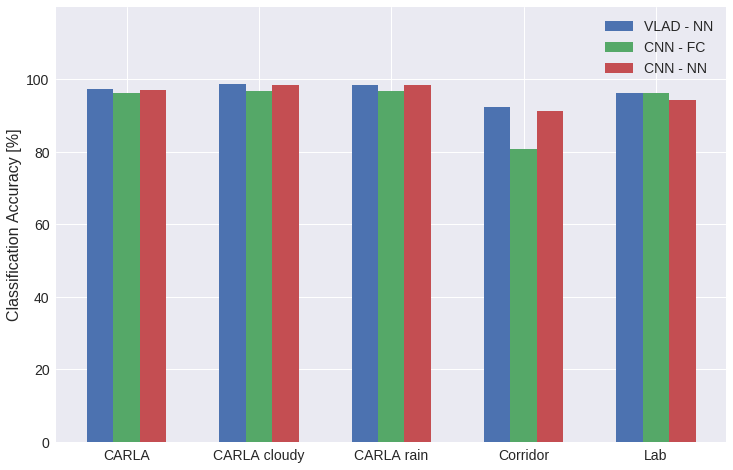}
    \caption{Topological accuracy across all datasets}
    \label{fig:topo_acc}
\end{figure}

We developed this technique at the Ford Palo Alto research centre, 
and hence, the labs and corridors of our office were the natural environment where we collected real data and tested our algorithm. We required multiple traversals of the same loop across time to allow accumulation of enough images per node and training of the CNN and hence, did not use the most common publicly available dataset KITTI, which contains relatively few loops. We also attempted to use the Oxford Robotcar Dataset~\cite{RobotCarDatasetIJRR}, as it contains several loops, but due to issues such as camera exposure problems, we could not get accurate positioning information from ORB-SLAM.
Our environment comprises large floor-to-ceiling windows and indoor ceiling lights. The brightly natural-lit windows present large, saturated regions in the image during daytime and dark areas at night. Some indoor lights go off during the day, while more come on during the evening and night. The heavily polished floors present confounding reflections in large regions of the images in our Corridor dataset. The data comprises of image sequences captured from the stereo Zed camera \cite{stereozed} mounted on a trolley with a laptop doing multiple loops around the Corridor that occupies an area of 25m x 18m, and the smaller Lab, that is of size 5m x 4m.
Our algorithm was tested 
during day and night and lights-on and lights-off conditions. In our experiment, we map the environment with footage taken during daytime and localize on data captured in the evening. Example images can be found in figure \ref{fig:light_corr}.
As our ground truth for these sequences, we use ORB-SLAM to map out the area with a stereo camera. While our experiments only use the left camera of the stereo sensor at test time, we use both sensors when mapping the environment and getting the `true' localizations of each camera frame.
Results are presented in Table \ref{tab:localizer_comparison}, the accuracy of the classifier is found in figure \ref{fig:topo_acc}. For our experiments in these environments, we use $D_{thresh}=2$ and $\lambda=2$.

\subsection{Selecting Number of Nodes}
\begin{figure}
    \centering
    \includegraphics[width=0.8\linewidth]{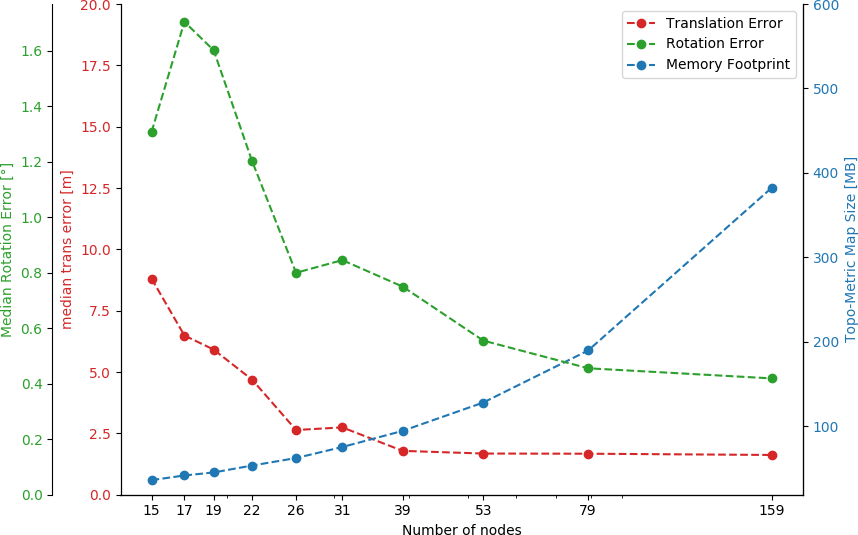}
    \caption{Effect of the number of nodes in the CARLA environment.}
    \label{fig:effectnodes}
\end{figure}

An important parameter of our method is setting the threshold distance $D_{thresh}$ at which a new node is placed (see Eq.~\ref{eq:thresh}). This parameter effectively determines the number of nodes that are created. In turn, this also determines the memory footprint of our map: more nodes, means more data to be stored. 
During our experiments, we found that this parameter is not very sensitive w.r.t. the error metrics and works for most reasonable values. To confirm this idea, we test this by starting from a very low threshold
distance, resulting in many nodes, and systematically increasing this threshold. The resulting error metrics, along with the memory footprint are plotted in figure~\ref{fig:effectnodes}. 

The rotation and translation errors decrease as more nodes are being added. Eventually, adding more nodes gives diminishing returns, and the error stops decreasing while the memory footprint rises. We believe that at this point the accuracy is limited by the accuracy of the fine localizer: the SURF matching and PnP solving procedure.

\subsection{Discussion and Future Work}

\begin{figure}
    \centering
    \includegraphics[width=0.85\linewidth]{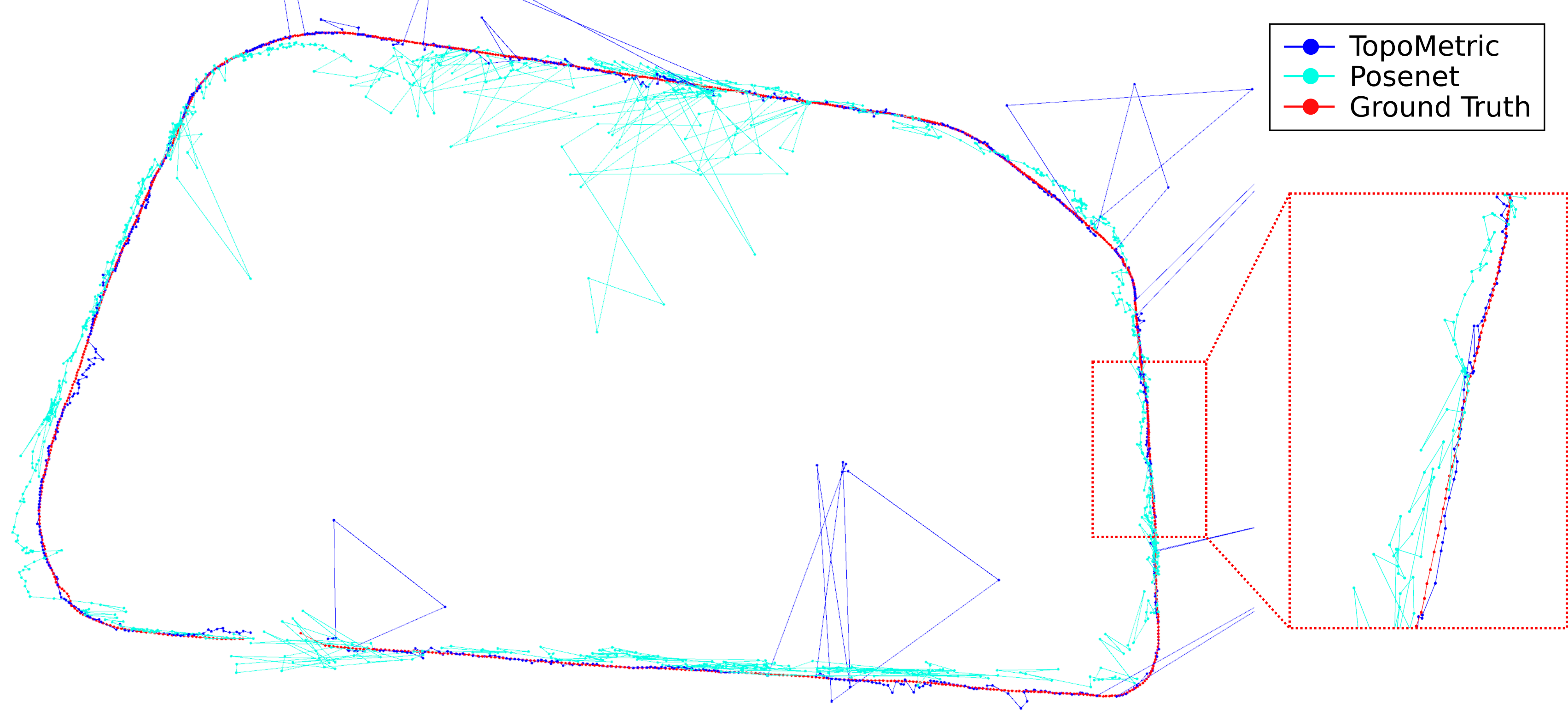}
    \caption{A qualitative comparison of Posenet  and our method (CNN - FC).}
    \label{fig:dl_vs_posenet}
    \vspace{-0.5cm}
\end{figure}



\begin{table}[]
\vspace{-0.5cm}
\caption{Average time spent on each aspect of our methods in seconds. The total includes overhead.}
\label{tab:time}
\centering
\begin{tabular}{@{}cccccccc@{}}
\toprule
     & CNN    & NN      & Patch Norm. & SURF+PNP & Total  \\ \midrule 
CNN-FC   & 0.0086 &  /      &  0.0020    & 0.0238   & 0.0553 \\
CNN-NN   & 0.0095 &  0.198  &  0.0020   & 0.0228   & 0.256 \\
VLAD-NN  & 0.0112 & 0.267   &  0.0020    & 0.0238   & 0.324   \\ \bottomrule
\end{tabular}
\end{table}

As a baseline, we apply a pure deep learning method, Posenet~\cite{kendall2015posenet}, to this problem. Posenet reported that they were able to outperform SIFT-based localizers. We train the network with the same ground truth position and orientation labels as our methods. In the case of the Lab and Corridor environments, these are the positions as found by ORBSLAM. We compare the quantitative results with our method in table \ref{tab:localizer_comparison} and see that it performs significantly worse. A qualitative example can be found in figure \ref{fig:dl_vs_posenet}, using one of the corridor sequences. Posenet shows some strong deviations from the ground truth path, much more frequently than our method.

The runtime of our proposed methods vary. Our fastest method is the \textit{CNN-FC}, as it is an end-to-end trained neural network running on a GPU. Our other methods are slower due to the complexity of our naive K-nearest neighbour classifier increasing with the number of training examples. A GPU implementation could make this much faster \cite{garcia_fast_2008}. A detailed account of the time spent on each subtask can be found in table \ref{tab:time}. This was measured on a machine with an Intel i7-4770K CPU and an NVIDIA GeForce GTX 1660.

Though the nearest neighbour based methods are more computationally expensive, they do not require any retraining of the CNN. Though it has been shown~\cite{arandjelovic_netvlad} that the VLAD descriptors are more robust to changes, in our experiments the performance of \textit{VLAD-NN} is comparable to \textit{CNN-NN}.

Our method is not without limitations. As shown in our experiment where we localized in an area in different weather conditions, our fine localizer fails. This can be attributed to a flawed keypoint descriptor and replacing it with one that is more robust to these changes would improve localization.

We have shown that we can determine the 6 DoF global pose of a camera in a topo-metric map using a combination of Deep Learning and geometric methods. Our map at the moment is made by the mapping stereo camera travelling in a loop in one direction only. To make this more applicable to real-world applications, we will need to include traversals in both directions. The number of outputs in the Topo-CNN will double, one matching the node for each direction of travel. Two sets of SURF features and their 3D coordinates will also need to be stored per topo-node. Junction nodes with more than two directions of travel might also need to be included in the topo-metric map for more complicated environments. 

At the moment, we use 2D image points and their SURF features to match a new image to a template image for a topo-node.  The long-term maintenance of these 2D feature points is more difficult compared to say larger, more permanent 3D structures in the scene. Indoor environments are Manhattan-esque, with lots of vertical lines, and using vertical lines or surfels occupying a larger area for feature matching and PnP could improve map maintenance over longer periods of time. Another interesting research question is how new features detected from monocular traversals at test time supplant old features that degrade over time.

Our matching is light-weight and uses only a single image at the moment. Every new image at test time is matched independently to the template features from the topo-node. Tracking of local features across test images to determine a local trajectory using VO could be used to bolster the single image PnP. A local optimization / bundle adjustment procedure using poses and points across images should improve pose estimation, at the cost of increased computational complexity.

These are all interesting directions of exploration that will be avenues of future research.

\section{CONCLUSIONS}
We propose the Deep-Geometric Localizer: a novel, monocular localization method utilizing the strengths of both deep learning and geometric computer vision. We show that it is possible to use a deep learning classifier to coarsely localize a camera inside a known environment and refine that position using keypoint matching and a PNP solver. Moreover, our method can localize in an environment up to absolute scale with a single camera, after mapping the area with a stereo camera. We tested our method in both simulated and real environments. We show that our proposed method outperforms Posenet, a state of the art CNN-based localization method that regresses 6 DoF pose from a single image. This suggests that there is much to gain by combining the fields of geometric computer vision and Deep Learning.

\addtolength{\textheight}{-12cm}   



\bibliographystyle{IEEEtran}
           
\bibliography{topoLoc}

\end{document}